# Neuromorphic Robust Framework for Concurrent Estimation and Control in Dynamical Systems using Spiking Neural Networks

Reza Ahmadvand, Sarah Safura Sharif, Yaser Mike Banad

*Abstract-* **Concurrent estimation and control of robotic systems remains an ongoing challenge, where controllers rely on data extracted from states/parameters riddled with uncertainties and noises. Framework suitability hinges on task complexity and computational constraints, demanding a balance between computational efficiency and mission-critical accuracy. This study leverages recent advancements in neuromorphic computing, particularly spiking neural networks (SNNs), for estimation and control applications. Our presented framework employs a recurrent network of leaky integrate-and-fire (LIF) neurons, mimicking a linear quadratic regulator (LQR) through a robust filtering strategy—modified sliding innovation filter (MSIF). Benefiting from both the robustness of MSIF and the computational efficiency of SNN, our framework customizes SNN weight matrices to match the desired system model without requiring training. Additionally, the network employs a biologically plausible firing rule similar to predictive coding. In the presence of uncertainties, we compare the SNN-LQR-MSIF with non-spiking LQR-MSIF and the optimal linear quadratic Gaussian (LQG) strategy. Evaluation across a workbench linear problem and a satellite rendezvous maneuver, implementing the Clohessy-Wiltshire (CW) model in space robotics, demonstrates that the SNN-LQR-MSIF achieves acceptable performance in computational efficiency, robustness, and accuracy. This positions it as a promising solution for addressing concurrent estimation and control challenges in dynamic systems.**

*Index Terms*: **Neuromorphic computing, Spiking neural network, Modified sliding innovation filter, Linear quadratic Gaussian, Satellite rendezvous maneuver.**

## I. INTRODUCTION

As the design and implementation of robotic manipulators/systems undertaking diverse real-world tasks grow more ambitious, the importance of computational efficiency, reliability, and accuracy escalates. Currently, all the implemented controllers rely heavily on the provision of accurate information about the system states/parameters obtained through various types of sensors, a task that often proves elusive due to the multifaceted uncertainties inherent to robotic systems. These uncertainties encompass environmental instability, unmodeled dynamics, and sensor noises, all of which can lead to data degradation, ultimately impacting controller performance. Furthermore, in some scenarios, obtaining complete measurements of all the states and parameters that describe the dynamics remains an impractical endeavor. Consequently, the ability to perform estimation simultaneously with control operations is paramount for ensuring the safe and accurate manipulation of robotic systems [1, 2].

In light of the constraints imposed by computing resources and energy consumption, the development of concurrent estimation and control frameworks that excel in computational efficiency, robustness, and accuracy becomes an imperative endeavor. The linear quadratic Gaussian (LQG) which is a popular and optimal framework for simultaneous estimation and control of linear dynamical systems, has found widespread adoption across various domains such as robotic manipulators [3], robot control [4], robot path planning [5], and satellite control [6]. However, the LQG framework is not without its limitations. The LQG framework is a linear quadratic regulator (LQR) that works based on the state feedback provided by the Kalman filter (KF) [7]. When confronted with uncertain dynamic models, its performance diminishes, and in the presence of external disturbances, it is not robust enough [8]. In such circumstances, the KF employed in conjunction with LQR control falls short of providing accurate information about system states/parameters. Consequently, the demonstrated limitations of the LQG underscore the pressing need for the development of a framework grounded in robust estimation principles.

In this study, we introduce a novel framework, LQR-MSIF, which combines the LQR controller with a recently introduced robust filtering strategy known as modified sliding innovation filter (MSIF) [9, 10]. The LQR-MSIF leverages the robustness of the MSIF filter in processing signals obtained from measurement systems. The MSIF represents an evolution of the sliding innovation filter (SIF), which belongs to the family of variable structure filters (VSF) [11], and also it can be considered as a new generation of smooth variable structure filter (SVSF) [12]. Importantly, unlike the KF family, which prioritizes frameworks founded on minimal estimation error, the VSF family of algorithms has been developed based on guaranteed stability in the presence of bounded modeling uncertainties and external disturbances [13].

Additionally, considering the recent advancements in neuromorphic computing tools, including spiking neural networks (SNN), and their applications in robotics control and estimation [10, 14], as well as the spike coding theories [15],

Reza Ahmadvand, Sarah Safura Sharif and Yaser Mike Banad (corresponding author) are with the Department of Electrical, and Computer Engineering, University of Oklahoma, Oklahoma, United States E-mails: *iamrezaahmadvand1@ou.edu, s.sh@ou.edu, bana@ou.edu*

we present a pioneering approach. In this study, to introduce a framework that comprehensively addresses the aforementioned limitations, we translate the LQR-MSIF into a neuromorphic SNN-based framework, in which the firing rule derived from the predicted error of the network concerning the estimated state vector, constituting a manifestation of predictive coding [15]. This theory posits that the brain perpetually constructs and enhances a 'mental model' of its surrounding environment, serving the critical function of anticipating sensory input signals, which are subsequently compared with the actual sensory inputs received. As the concept of representation learning gains increasing prominence, predictive coding theory has found vibrant application and exploration within the realms of biologically inspired neural networks, such as SNN. The adoption of SNNs mitigates the computational efficiency challenges associated with this problem [16]. Owing to their minimal computational burden and inherent scalability, SNNs offer significant advantages over traditional non-spiking computing methods [17].

SNNs represent the third generation of neural networks, taking inspiration from the human brain, where neurons communicate using electrical pulses called spikes. SNNs leverage neural circuits composed of neurons and synapses, communicating via encoded data through spikes in an asynchronous fashion [17, 18, 19, 20, 21]. The asynchronous in spiking fashion characterized by event-driven processing [10], stands in contrast to traditional Artificial Neural Networks (ANNs), which operate synchronously or, in other words, are time-driven. Studies [22] demonstrate that, for equivalent tasks, SNNs are 6 to 8 times more energy efficient than ANNs with an acceptable trade-off in accuracy [23]. Moreover, the inherent scalability of SNNs enhances their reliability, particularly under the condition of neuron silencing, where neuron loss is compensated for by an increase in the spiking rate of remaining neurons [18].

Thus, to harness the advantage of SNNs for the simultaneous robust estimation and control, here, we integrate the methods proposed in prior studies [10], and [14] to develop the previously mentioned SNN-LQR-MSIF framework, anticipating substantial advantages. Subsequently, we assess the performance of the proposed SNN-LQR-MSIF framework through a series of evaluations. Initially, we apply it to a linear workbench problem, followed by its application to the intricate task of satellite rendezvous in circular orbit, a critical maneuver in space robotic applications such as on-orbit servicing and refueling [24], We then compare the SNN-LQR-MSIF with its non-spiking counterpart, LQR-MSIF, and the standard LQG under various sources of uncertainty, including modeling uncertainty, measurement outliers, and neuron silencing. For the proposed framework, our findings revealed an acceptable performance in terms of curacy, and robustness while it outperforms the traditional frameworks in terms of computational efficiency

This paper is organized as follows. Section 2 provides an overview of related works and contributions. Then, the preliminaries, underlying theories, and the proposed framework for addressing the problem of concurrent robust estimation and control in linear dynamical systems are presented in Section 3. Next, Section 4 provides numerical simulations and discussions of the results, while Section 5 serves as the conclusion of the paper.

## II. RELATED WORKS AND CONTRIBUTIONS

In this section, an overview of recent related works, and our contributions have been presented separately.

### A. Related works

This section offers a concise overview of recent works related to the problem of concurrent estimation and control. In [14], Yamazaki *et al*, proposed an SNN-based framework for concurrent estimation and control, employing a combination of the Luenberger observer and LQR controller. They applied their method to scenarios involving a spring-mass-damper (SMD) system and a Cartpole system, evaluating its performance in terms of accuracy and similarity to its non-spiking counterpart. They also explored the robustness of their network in handling neuron silencing. While their results were promising, their framework had limitations, notably the need to design both controller and observer gains for each problem. Additionally, since they used the Luenberger observer, their framework inherited the observer limitations related to modeling uncertainties and external disturbances, which were not thoroughly assessed for robustness.

To address these limitations, a novel SNN-based KF was proposed in [10] for optimal estimation of linear dynamical systems. In addition to performing the optimal estimation, this approach eliminated the need for observer gain design, simplifying the process. To enhance robustness against modeling uncertainties and external disturbances, a robust SNN-based estimation framework based on MSIF was introduced. Comparative assessments involving traditional KF and MSIF demonstrated acceptable performance for the SNN-based frameworks in terms of similarity to non-spiking strategies, robustness, and accuracy. However, the previous study did not investigate concurrent estimation and control scenarios, which is the primary focus of this research. Additionally, none of the aforementioned methods utilized biologically inspired firing rules for their network.

### B. Contributions

The contributions of our research are as follows:

- **Development of SNN-LQR-MSIF:** We introduce a robust SNN-based framework for concurrent estimation and control of linear dynamical systems, named SNN-LQR-MSIF. This framework leverages previously proposed methods in [10] and [14].
- **Biologically Plausible Firing Rule:** In order to have control over the spike distribution in the network and prevent excessive spiking for a part of the network or a neuron, we implement a biologically plausible firing rule based on the concept of predictive coding concept [15], enhancing the biological relevance of our network.
- **Robustness and Accuracy Assessment:** We comprehensively investigate the performance of our method in scenarios subjected to modeling uncertainties, measurement outliers, and neuron silencing, evaluating robustness and accuracy compared to its non-spiking counterpart LQR-MSIF and the traditional LQG. We also analyze spiking patterns to demonstrate computational efficiency.
- **Application to Satellite Rendezvous:** We apply the SNN-LQR-MSIF to a real-world scenario involving concurrent estimation and control of satellite rendezvous, a novel application for this type of neuromorphic framework. We compare its performance with that of LQR-MSIF and LQG.

## III. Theory

In this section, we provide essential preliminaries, followed by an outline of the study's outcomes. The linear dynamical system and measurement package considered in this study are defined by the following equations:

$$\dot{x} = Ax + Bu + w \tag{1}$$
$$z = Cx + d \tag{2}$$

Here, $x \in R^{n_x}$ refers to the state vector, $u \in R^{n_u}$ is the input vector, $z \in R^{n_z}$ is the measurement vector. $A \in R^{n_x \times n_x}$, and $B \in R^{n_x \times n_u}$ denote the dynamic transition and input matrices, respectively, while $C \in R^{n_z \times n_x}$ is the measurement matrix. $w$ and $d$ represent the zero-mean Gaussian white noise with covariance matrices $Q$, and $R$, respectively.

Figure 1 depicts the traditional block diagram of a concurrent estimation and control loop in conventional dynamical systems. This diagram reveals that both the estimator and controller employ sequential algorithms, resembling the logic of traditional von Neumann computer architectures.

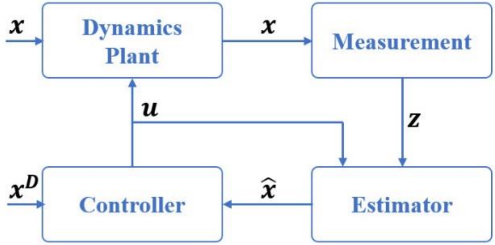

Fig. 1. Conventional block diagram of concurrent estimation and control loop in traditional dynamical systems

*A. Spiking neural networks (SNN)*

In this section, we present a brief overview of implementing an SNN, including its firing rule. To design a network composed of recurrent leaky integrate-and-fire (LIF) neurons capable of approximating the temporal variation of a parameter like $x$ as expressed in Eq. (1), we need to implement the following equation [14]:

$$\dot{\sigma} = -\lambda \sigma + D^T(\dot{x} + \lambda x) - D^T D s \tag{3}$$

Here, $\sigma \in R^N$ refers to the neuron membrane potential vector, $\lambda$ is a decay or leak term considered on the membrane potential of the neurons, $D \in R^{n_x \times N}$ is the random fixed decoding matrix containing the neurons' output kernel, and $s \in R^N$ is the emitted spike population of the neurons in each time step. Further, according to spike coding network theories [14, 15], the introduced network of LIF neurons can reproduce the temporal variation of $x$ under two assumptions. First, we should be able to estimate $x$ from neural activity using the following rule:

$$\hat{x} = Dr \tag{4}$$

Here, $r \in R^N$ represents the filtered spike trains, which have slower dynamics compared to $s \in R^N$. The dynamics of the filtered spike trains are provided by:

$$\dot{r} = -\lambda r + s \tag{5}$$

The second assumption is that the network minimizes the cumulative error between the true value of $x$ and the estimated $\hat{x}$, leveraging optimization on the spike times not by changing the output kernel values $D$. So, the network minimizes the cumulative error between the state and its estimate while limiting computational cost by controlling spike occurrence. To achieve this, it minimizes the following cost function [15]:

$$J = \int_0^t (\|x(\tau) - \hat{x}(\tau)\|_2^2 + \nu\|r(\tau)\|_1 + \mu\|r(\tau)\|_2^2) d\tau \tag{6}$$

Here, $\|.\|_2^2$ represents the Euclidean norm, and $\|.\|_1$ indicates L1-norm. This firing rule ensures that each neuron emits a spike only when it contributes to reducing the predicted error, resembling a form of predictive coding [15].

Finally, along with obeying the mentioned rules, the neurons will emit spikes when their membrane potential reaches their specific thresholds. Thus, to ensure a biologically plausible spiking pattern, we define thresholds using the following expression:

$$T_i = \frac{D_i^T D_i + \nu\lambda + \mu\lambda^2}{2} \tag{7}$$

Here, $D_i$ is $i^{th}$ column of the matrix $D$, which represents $i^{th}$ neuron's output kernel that reflects the change in the error due to a spike of $i^{th}$ neuron. The parameters $\nu$ and $\mu$ control the trade-off between computational cost and accuracy, with $\nu$ encouraging the network to use as few spikes as possible, while the quadratic term influenced by $\mu$, which is responsible for a uniform distribution of spikes among network neurons. Proper tuning of these parameters yields biologically plausible spiking patterns where neural activity is distributed approximately evenly among neurons. This firing rule operates based on the concept that the network minimizes the error by controlling spike occurrence in response to excitation and inhabitation, ultimately reducing the prediction error a kind of predictive coding.

*B. SNN-based robust filtering*

The robust SNN-based filtering strategy for linear dynamical systems referred to as SNN-MISF, was previously introduced in [10]. SNN-MSIF combines the SNN with the MSIF, a robust filtering strategy for dynamical systems [9]. In this framework, SNN-MSIF is represented as a recurrent SNN composed of leaky integrate-and-fire (LIF) neurons. Its weight matrices can adaptively change to emulate the linear dynamics of the MSIF, combining the computational efficiency and scalability of

SNNs with the robustness of the MSIF. The equations governing SNN-MSIF are as follows [10]:

$$\dot{\boldsymbol{\sigma}} = -\lambda\boldsymbol{\sigma} + F\boldsymbol{u}(t) + \Omega_s \boldsymbol{r} + \Omega_f \boldsymbol{s} \\ + \Omega_k \boldsymbol{r} + F_k \boldsymbol{z} + \boldsymbol{\eta} \quad (8)$$

where:

$$F = D^T B \quad (9)$$
$$\Omega_s = D^T (A + \lambda I) D \quad (10)$$
$$\Omega_f = -(D^T D + \mu\lambda^2 I) \quad (11)$$

Here, $\lambda$ represents the leak rate for the membrane potential, and $F$ encodes the control input to a set of spikes that is readable for the network. $\Omega_s$ and $\Omega_f$ are synaptic weights for slow and fast connections, respectively. While slow connections typically govern the implementation of the desired system dynamics, in this context, they are chiefly responsible for executing the linear dynamics of the MSIF estimator. Conversely, fast connections play a pivotal role in achieving an even distribution of the spikes across the network. Consequently, the primary contributors to the *a-priori* prediction phase of the estimation process are the second three terms in Eq. (3). In contrast, the subsequent two terms, which are influenced by $\Omega_k$, and $F_k$, adapt dynamically during the estimation process, and are tasked with handling the measurement-update or *a-posteriori* phase of the estimation. Here, $\Omega_k$ imparts the dynamics of the update component, while $F_k$ furnishes the SNN with an encoded measurement vector. To update these weight matrices the following expressions, need to be used:

$$\Omega_k = -D^T (C^+ sat(diag(P^{zz})/\delta)) CD \quad (12)$$
$$F_k = D^T (C^+ sat(diag(P^{zz})/\delta)) \quad (13)$$

Here, $P^{zz}$ represents the innovation covariance matrix, and $\delta$ is the sliding boundary layer, a tuning parameter. To update $P^{zz}$, the following equations are used:

$$P^{zz} = PCP^T + R \quad (14)$$
$$\dot{P} = AP + PA^T + Q - PC^T R^{-1} CP \quad (15)$$

The final term $\boldsymbol{\eta}$, accounts for zero-mean Gaussian noise, simulating the stochastic nature of the neural activity in biological neural circuits. The weight matrices are analytically designed to capture MSIF dynamics, allowing the estimation of a fully observable linear dynamical system with partially noisy state measurements via a network of recurrent LIF neurons.

Utilizing the framework presented in this section for concurrent estimation concurrently with the conventional control methods results in the system depicted in Fig. 2. The figure illustrates how the conventional non-spiking estimator in Fig. 1 has been replaced by an SNN designed to function as an estimator. Instead of employing sequential estimation algorithms, this SNN-based approach capitalizes on the advantages of SNNs, including computational efficiency, highly parallel computing, and scalability. However, as shown in Fig. 2, estimation and control tasks are still conducted sequentially.

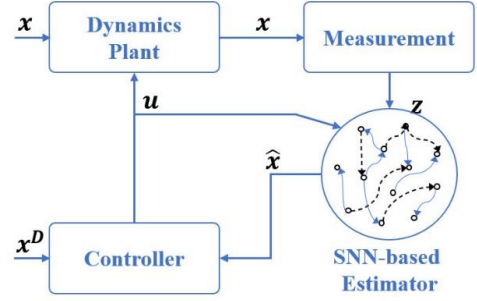

Fig. 2. Conventional block diagram of concurrent estimation and control loop of dynamical systems using SNN-based estimator

*C. SNN-based concurrent estimation and control*

This section extends SNN-MSIF to a network capable of concurrently performing state estimation and control of linear dynamical systems. As introduced in [10], for the derivation of the SNN-MSIF, which implements the linear dynamics of an estimator, the SNN should be able to mimic the following dynamics:

$$\dot{\hat{\boldsymbol{x}}} = A\hat{\boldsymbol{x}} - B\boldsymbol{u} + K_{KF}(\boldsymbol{z} - \hat{\boldsymbol{z}}) \quad (16)$$

Here, to go further and add the control to the above-mentioned dynamics; $\boldsymbol{u} = -K_c(\boldsymbol{x} - \boldsymbol{x}^D)$ is considered as the control input, So, the network should emulate the following linear system of equations:

$$\dot{\hat{\boldsymbol{x}}} = A\hat{\boldsymbol{x}} - BK_c(\hat{\boldsymbol{x}} - \boldsymbol{x}^D) + K_{KF}(\boldsymbol{z} - \hat{\boldsymbol{z}}) \quad (17)$$

where $\boldsymbol{x}^D$ denotes the desired state. To extend the previously introduced network, the control rule $\boldsymbol{u}$ is substituted into Eq. (8), resulting in the following network equation:

$$\dot{\boldsymbol{\sigma}} = -\lambda\boldsymbol{\sigma} - FK_c(\hat{\boldsymbol{x}} - \boldsymbol{x}^D) + \Omega_s \boldsymbol{r} + \Omega_f \boldsymbol{s} \\ + \Omega_k \boldsymbol{r} + F_k \boldsymbol{z} + \boldsymbol{\eta} \quad (18)$$

Here, a decoding rule for the $\boldsymbol{x}^D$ is considered as follows:

$$\boldsymbol{x}^D = \bar{D}\boldsymbol{r} \quad (19)$$

where $\bar{D}$ is a fixed matrix generated from random elements sampled from a zero-mean Gaussian distribution. To effectively implement the added dynamics for $\boldsymbol{x}^D$ in the network, referring to Eq. (3), an additional set of connections is introduced into the above equation [14]:

$$\dot{\boldsymbol{\sigma}} = -\lambda\boldsymbol{\sigma} - D^T BK_c (D\boldsymbol{r} - \bar{D}\boldsymbol{r}) + \Omega_s \boldsymbol{r} \\ + \Omega_f \boldsymbol{s} + \Omega_k \boldsymbol{r} + F_k \boldsymbol{z} + \bar{D}^T (\dot{\boldsymbol{x}}^D + \lambda \boldsymbol{x}^D) \\ + \bar{\Omega}_f \boldsymbol{s} + \boldsymbol{\eta} \quad (20)$$

After simplifications, the resulting network equation is as follows:

$$\dot{\boldsymbol{\sigma}} = -\lambda\boldsymbol{\sigma} + \Omega_c \boldsymbol{r} + \bar{\Omega}\boldsymbol{r} + \Omega_s \boldsymbol{r} + \Omega_f \boldsymbol{s} \\ + \Omega_k \boldsymbol{r} + F_k \boldsymbol{z} + \bar{D}^T (\dot{\boldsymbol{x}}^D + \lambda \boldsymbol{x}^D) + \bar{\Omega}_f \boldsymbol{s} \quad (21) \\ + \boldsymbol{\eta}$$

where:
$$\Omega_c = -D^T B K_c D \quad (22)$$
$$\bar{\Omega} = D^T B K_c \bar{D} \quad (23)$$
$$\bar{\Omega}_f = -(\bar{D}^T \bar{D} + \mu \lambda^2 I) \quad (24)$$

Here, $\Omega_c$ somehow represents the slow connections for implementing the control input of the desired system. $\bar{\Omega}$, and $\bar{\Omega}_f$ represent the slow and fast synaptic weights for various connections respectively. parallel with other connections, these weights are responsible for implementing the dynamics of the desired state for the controller and Eq. (18) represents the membrane potential dynamics of a recurrent SNN of LIF neurons, capable of concurrently performing state estimation and control of linear dynamical systems. While the controller gain $K_c$ must be designed for the considered system, this framework operates without requiring any learning by the network. Furthermore, although we implemented optimal LQR control in this study, the controller gain can be independently designed using any arbitrary approach. Finally, to extract the control input vector for the external plant from the spike populations, the following equation is employed:

$$\boldsymbol{u} = D_u \boldsymbol{r} \quad (25)$$
where:
$$D_u = -K_c (D - \bar{D}) \quad (26)$$

The above matrix can be used for decoding the control input from the neural activity inside the network. In summary, the proposed framework concurrently estimates the state vector $\boldsymbol{x}$ from a noisy partial measurement vector $\boldsymbol{z}$ and provides control input for the considered system. Fig. 3 illustrates the block diagram of the framework presented in this section.

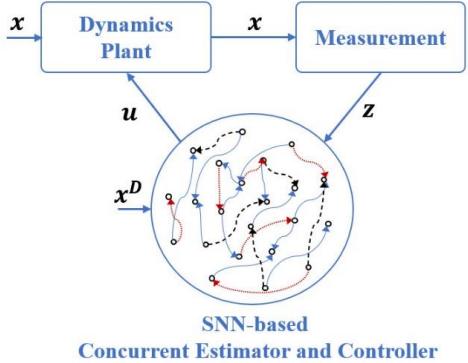

Fig. 3. Block diagram of SNN-based concurrent estimation and control loop.

Fig. 3 demonstrates that for this framework, both the blocks of estimator and controller from Fig. 1 and Fig. 2 have been replaced by a single SNN. This represents an extension of the framework, leveraging the advantages of SNNs. Furthermore, the computations required for state estimation and control input have been parallelized. Consequently, implementing this framework can significantly reduce computational costs, allowing more complex tasks to be performed even with limited computing resources. Additionally, owing to the scalability of SNNs, if a part of the implemented network becomes damaged or loses some neurons, the process continues by increasing the spiking rate of the remaining neurons, as demonstrated in the next section.

## IV. NUMERICAL SIMULATIONS

In this section, we first apply the proposed framework to a linear workbench problem and conduct various performance evaluations in terms of robustness, accuracy, and computational efficiency, in comparison with the well-established methods LQG and LQR-MSIF. Subsequently, we extend the analysis of the SNN-LQR-MSIF to a practical scenario involving the concurrent estimation and control of satellite rendezvous maneuvers.

### A. Case study 1: Linear workbench problem

Here, we initiate our investigation by applying the introduced framework to the following linear dynamical system:

$$\dot{\boldsymbol{x}} = \begin{bmatrix} 0 & 0 \\ 0 & 1 \end{bmatrix} \boldsymbol{x} + \begin{bmatrix} 0 \\ 1 \end{bmatrix} \boldsymbol{u} + \boldsymbol{w} \quad (27)$$
$$\boldsymbol{z} = [1 \quad 0] \boldsymbol{x} + \boldsymbol{v} \quad (28)$$
where:
$$\boldsymbol{u} = -K_c \boldsymbol{x} \quad (29)$$

Simulations have been performed over a 10-second period with a time step of 0.01, employing the numerical values provided in TABLE 1.

TABLE 1
LINEAR SYSTEM SIMULATION PARAMETERS

| Parameter | Value |
|---|---|
| $x_0$ | [10,1] |
| $\hat{x}_0$ | [10,1] |
| $K_c$ | [1, 1.7321] |
| $Q_c$ | $I$ |
| $R_c$ | $I$ |
| $Q$ | $I/1000$ |
| $R$ | $I/100$ |
| $N$ | 250 |
| $\lambda$ | 0.01 |
| $\mu$ | 0.005 |
| $\nu$ | 0.005 |
| $\delta_{MSIF}$ | 0.005 |

Initially, we evaluated the applicability of the proposed framework in comparison with its non-spiking counterparts, LQG, and LQR-MSIF, by simulating a deterministic system without uncertainties. Next, we assessed the performance and effectiveness of the proposed framework by introducing various sources of uncertainties and disturbances. In line with real-world scenarios, where exact decoding matrices are typically unknown, we defined the decoding matrices $D$ and $\bar{D}$ using random samples from zero-mean Gaussian distributions with covariances of 0.25 and 1/300, respectively.

Fig. 4 displays time histories of controlled states and estimation errors within $\pm 3\sigma$ bounds obtained from SNN-LQR-SIF in comparison with LQG, and LQR-MSIF. Fig. 4(a) illustrates that the state $x_1$ converges to zero after $t = 5$s, showcasing similar performance between the proposed framework and its non-spiking counterparts, LQG and LQR-MSIF. Fig. 4(b) indicates that the state $x_2$ converges to zero around $t = 6$s, again showing consistent performance between the proposed framework and non-spiking methods. Fig. 4(c) demonstrates that all considered strategies remain stable, with errors staying within the prescribed bounds. Notably, the error obtained from KF deviates further from zero before converging around $t = 3$s, while the errors from SNN-MSIF and MSIF exhibit faster convergence with smaller deviations. Fig. 4(d) confirms the stability of all estimation methods, with SNN-LQR-MSIF showing nearly identical performance to non-spiking KF and MSIF.

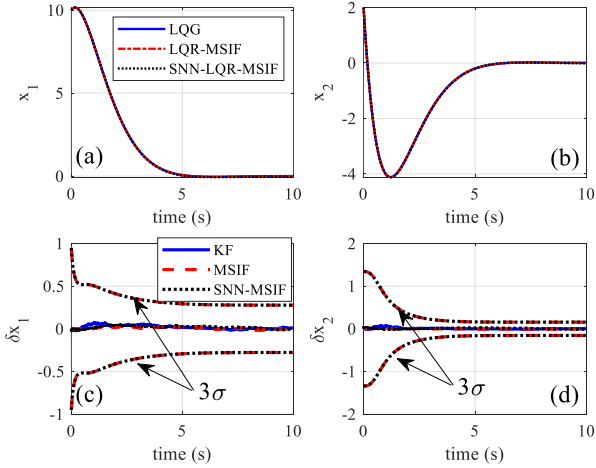

Fig. 4. Controlled states and estimation errors within $\pm 3\sigma$ bounds (a) controlled state $x_1$, (b) controlled state $x_2$, (c) estimation error of $x_1$, (d) estimation error of $x_2$

Further, to gain more intuitive insights into the tuning parameters of the firing rule, namely $\mu$, and $\nu$ and their impacts on control accuracy, we conducted a sensitivity analysis. as depicted in Fig. 5, utilizing a colored map to show the variations of normalized average error, this analysis reveals that parameters tuning directly affects control accuracy, and depending on the specific system proper parameter sets can be identified by trial and error. The preferred parameter set used throughout our simulations is $\mu = 0.005$ and $\nu = 0.005$ marked with a white circle in the figure. The percentage of emitted spikes by the neurons compared to all possible spikes is also shown in the figure by a number on the figure for each set of $\mu$ and $\nu$. It can be observed that decreasing $\nu$ leads to a higher percentage of spikes compared to possible spikes for each $\mu$. This highlights a trade-off between accuracy and computational efficiency that can be an important factor in the tuning procedure of the network firing rule and confirms the previously mentioned matter about the tuning of $\nu$ that controls the number of spikes.

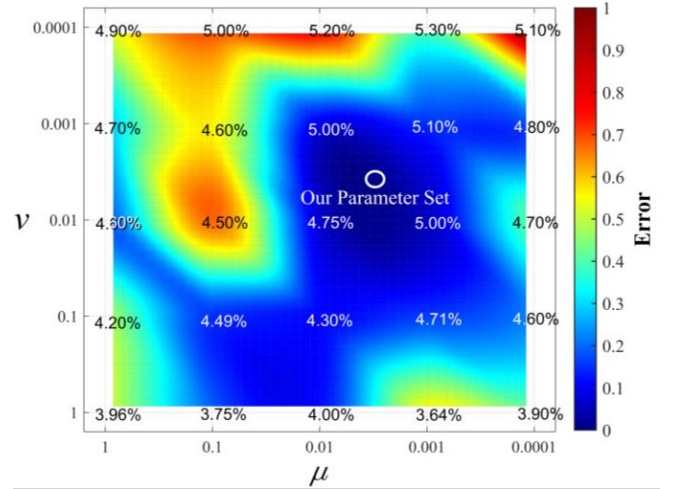

Fig. 5. Colored map analysis of normalized average error obtained from various sets of $\mu$ and $\nu$. Additionally, compared to all possible spikes for each set of $\mu$ and $\nu$ the number of emitted spikes in percent is presented.

Furthermore, we evaluated the robustness of SNN-LQR-MSIF against modeling uncertainties by introducing a 20% error in the dynamic transition matrix $\hat{A} = 0.8A$. Simulation results in the presence of modeling uncertainty were compared with LQG and LQR-MSIF, as presented in Fig. 6. Fig. 6(a) shows that in the presence of uncertainty, the SNN-based framework for the state $x_1$ deviates from non-spiking LQG and LQR-MSIF. However, SNN-LQR-MSIF exhibits superior performance, converging to zero at approximately $t = 4$s and completely converging by $t = 6$s. In contrast, non-spiking frameworks yield matching results converging to zero at $t = 7$s. Fig. 6(b) demonstrates that state $x_2$ exhibit similar deviation from non-spiking methods, particularly with a slightly greater overshoot and error until $t = 4$s. However, after $t = 4$s, SNN-LQR-MSIF displays faster convergence, a minor overshoot, and eventual convergence to zero after $t = 8$s. In summary, these findings indicate that the proposed SNN-based framework exhibits commendable robustness in handling modeling uncertainties or external disturbances compared to non-spiking methods. Fig. 6(c) illustrates the results for the state $x_1$, showcasing the performance of SNN-LQR-MSIF comparable to that of LQR-MSIF. Initially, both methods exhibit an error trend that diverges over time, exceeding the bound around $t = 1.5$s but returning within the bound by $t = 4$s. Eventually, both methods achieve stable estimation, converging to zero around $t = 6$s and $t = 8$s for SNN-MSIF and MSIF, respectively. Meanwhile, the error from KF deviates entirely and its error has returned to the bound in almost $t = 8$s and finally, it converged to zero at $t = 10$s. Notably, at $t = 6s$, KF exhibits an error that is approximately 20 times greater than the error obtained for the proposed SNN-LQR-MSIF is almost near to zero. In Fig. 6(d), the results for the state $x_2$ show nearly identical performance between SNN-MSIF and MSIF, both maintaining stability in their estimations throughout the considered period. Conversely, the error from KF deviates similarly to what occurred with the state $x_1$. The obtained error for KF has exceeded the bound and has risen continually until almost $t = 2.5$s reaches its maximum that is about 102 times greater than the obtained error for MSIF and SNN-MSIF is also approximately near to zero. Hence, it is evident that SNN-MSIF outperforms MSIF by faster convergence to zero in the presence of uncertainty, and it outperforms KF in terms of estimation stability.

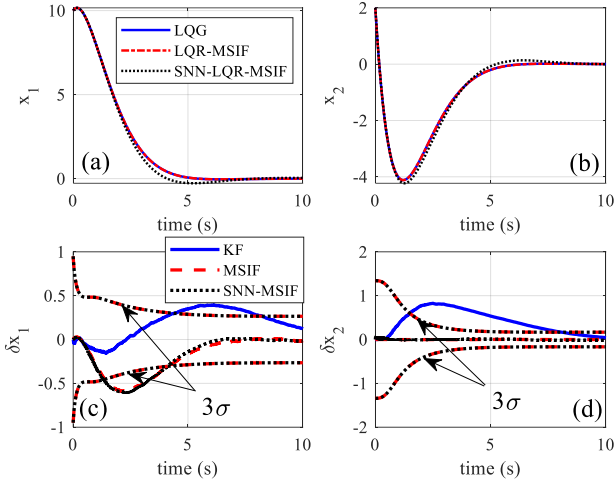

Fig. 6. Controlled states and estimation errors within $\pm 3\sigma$ bounds for uncertain model $\hat{A} = 0.8A$, (a) controlled state $x_1$, (b) controlled state $x_2$, (c) estimation error of $x_1$, (d) estimation error of $x_2$

An important challenge in robust navigation and control systems is handling measurement outliers, which can arise from sensor faults or external disturbances in the working environment. Therefore, to assess the framework's robustness in such scenarios, unmodeled measurement outliers were introduced into the system at $t = 3$s, $t = 5$s, and $t = 6$s. To simulate the presence of measurement outliers, the measurement system noise was multiplied by a factor of 500 at these time points. Fig. 7 presents a comparison of results for controlled states and estimation errors within $\pm 3\sigma$ bounds obtained from various frameworks in the presence of measurement outliers. Fig. 7(a) displays the time history of the state $x_1$. It demonstrates that the presence of measurement outliers causes slight deviations in the results obtained from the SNN-based framework between $t = 3$s, and $t = 7$s. However, the framework successfully regulates the error, ultimately converging to results obtained from non-spiking methods. Fig. 7(b) demonstrates the same behavior for the state $x_2$. Results from the SNN-based framework show minor deviations compared to non-spiking methods between $t = 3$s, and $t = 7$s, indicating that, although more sensitive to measurement outliers, the SNN-based methods continue to control the states effectively. Fig. 7(c) presents the obtained errors for the state $x_1$, which exhibit significant deviations at the points of outlier injection. However, for all considered filters, these deviations are followed by rapid convergence to zero, confirming the filters' stability. Moreover, the error from SNN-MSIF is considerably smaller, especially compared to KF which exceeds the bound on all points. In Fig. 7(d), we investigate the error for the state $x_2$ which reveals when KF experiences abrupt deviation and its error exceeds the bound at the points of outlier injection, whereas SNN-MSIF and MSIF remain stable throughout the simulation. Thus, SNN-MSIF exhibits superior robustness in such situations.

Fig. 8 illustrates the spiking pattern of the network achieved by the SNN-LQR-MSIF approach when confronted with measurement outliers. In Fig. 8(a), we present the spiking pattern recorded in the presence of measurement outliers. It is evident that just right before the points of outlier injections (at time steps 300, 400, and 600), most neurons are in standby mode, emitting a few spikes. However, after the introduction of outliers, a substantial portion of neurons (around 40%) become activated to handle the injected disturbances, which are rejected within just 2-3 time steps. The neural activity then decreases, demonstrating that the network effectively overcomes external disturbances or unmodeled dynamics by increasing neural activity or computational cost without failing in the assigned task. Moreover, Fig. 8(b) reveals the temporal variation of active neurons in percent, emphasizing the sudden change in the population of active neurons at the designated time steps. The population rises to nearly 40% to overcome the negative impacts of injected outliers on the system.

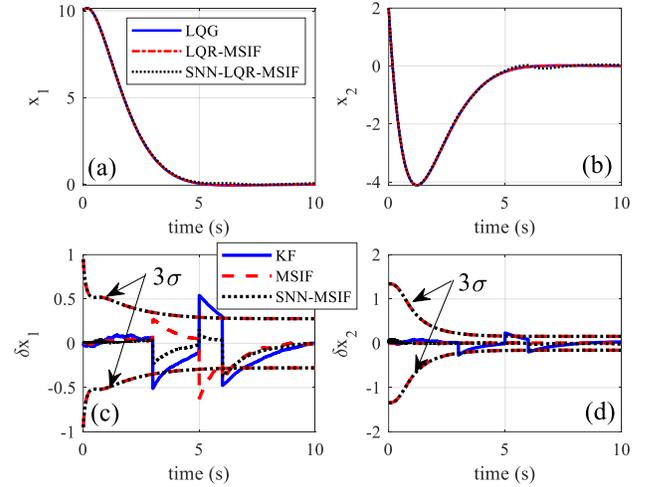

Fig. 7. Controlled states and estimation errors within $\pm 3\sigma$ bounds for measurement outlier (a) controlled state $x_1$, (b) controlled state $x_2$, (c) estimation error of $x_1$, (d) estimation error of $x_2$

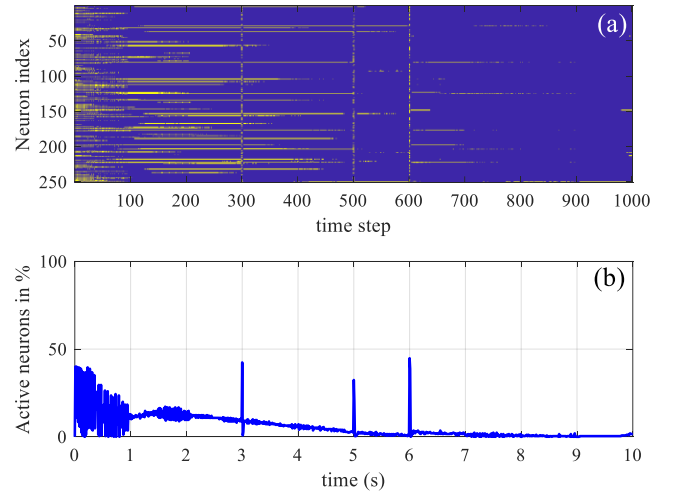

Fig. 8. Spiking pattern and temporal variation of active neurons population obtained from SNN-LQR-MSIF, (a) spiking pattern, (b) temporal variation of active neurons

Finally, to assess the proposed framework's performance in situations where some neurons may become silent, several simulations were conducted with varying numbers of neurons, ranging from $N = 50$ to $N = 400$ in the step of 50 neurons. Fig. 9 presents the average overall network error in the controlled states after $t = 6$s (where the errors almost converged to zero) versus the number of neurons. In region 1, a significant error divergence to infinity is observed (the solid line which shows the error variation became almost vertical at the edge of region 1) while this error is abruptly decreased at $N = 100$. This corresponds to the minimum number of neurons that the proposed framework requires to function effectively. Below this threshold, active neurons cannot provide sufficient neural activity to perform the necessary computations. An increase in the number of neurons within region 2 results in a gentle reduction in error. The minimum error can be observed at the optimal number of neurons at $N = 250$. In contrast, region 3 shows that an increase in the number of neurons degrades accuracy due to unstable spiking patterns with excessive neural activity.

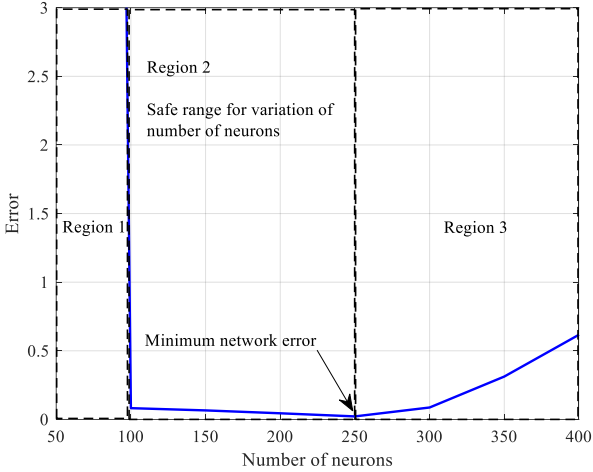

Fig. 9. Averaged network error versus number of neurons (because of the huge divergence of error in region 1, the solid line became almost vertical at the edge of region 1)

Overall, the proposed framework exhibits remarkable robustness in handling measurement outliers and effectively adapts to situations with varying numbers of neurons, provided a minimum neuron threshold is maintained. These findings support the framework's suitability for robust navigation and control systems in real-world scenarios. Further studies on spiking patterns are provided in [10].

*B. Case study 2: Satellite rendezvous maneuver*

This section is initiated by the presentation of the mathematical model for the satellite rendezvous maneuver. Subsequently, the design of the LQR controller is expounded upon. Lastly, the simulation results are provided. The rendezvous problem involves maneuvering two distinct satellites, the chaser, and the target. As depicted in Fig. 10, the chaser satellite approaches the target in orbit.

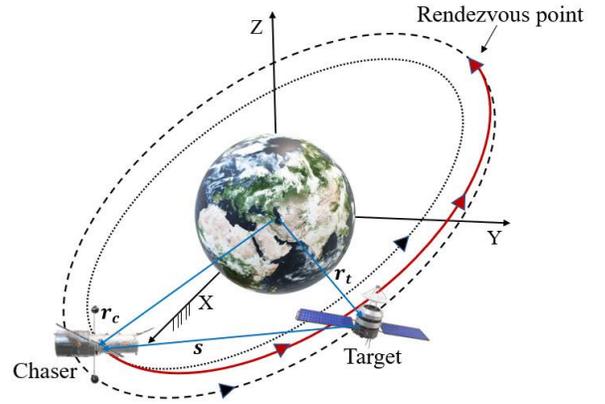

Fig. 10. Schematic of rendezvous maneuver

To derive the equations of relative motion, we consider the following equation in the Earth-centered inertial frame (ECI) [25].

$$\boldsymbol{s} = \boldsymbol{r}_c - \boldsymbol{r}_t \qquad (30)$$

Here, $\boldsymbol{r}_c$ and $\boldsymbol{r}_t$ represent the position vectors of the chaser and target, respectively. The relative acceleration is described by the following expression:

$$\ddot{\boldsymbol{s}} = \ddot{\boldsymbol{r}}_c - \ddot{\boldsymbol{r}}_t \qquad (31)$$

Meanwhile, considering the circular orbit, the gravitational force in ECI is expressed as:

$$f_g(\boldsymbol{r}) = -\mu_{earth} \frac{m}{r^3} \boldsymbol{r} \qquad (32)$$

Here, $\mu_{earth}$ signifies the Earth's gravitational parameter, $m$ denotes spacecraft mass, and $\boldsymbol{r}$, and $r$ represent the spacecraft position vector and its magnitude, respectively. Importantly, the absolute motion of both the chaser and target in the ECI frame can be separately formulated as follows:

$$f_g(\boldsymbol{r}_t) = \ddot{\boldsymbol{r}}_t = -\frac{\mu_{earth}}{r_t^3} \boldsymbol{r}_t \qquad (33)$$

$$f_g(\boldsymbol{r}_c) = \ddot{\boldsymbol{r}}_c = -\frac{\mu_{earth}}{r_c^3} \boldsymbol{r}_c \qquad (34)$$

The above equations represent normalized forms of Eq. (32), divided by the spacecraft mass. To formulate suitable equations for controller design, it is advantageous to represent relative motion in the target frame, a non-inertial reference frame rotating with the angular velocity, $\boldsymbol{\omega}$.

$$\frac{d^{*2}\boldsymbol{s}^*}{dt^2} + \boldsymbol{\omega} \times (\boldsymbol{\omega} \times \boldsymbol{s}) + 2\boldsymbol{\omega} \times \frac{d^*\boldsymbol{s}^*}{dt} \\ + \frac{d\boldsymbol{\omega}}{dt} \times \boldsymbol{s}^* + \frac{\mu_{earth}}{r^3} M \boldsymbol{s}^* = \boldsymbol{f} \qquad (35)$$

Here, $\boldsymbol{s}$ denotes relative distance, $M$, and $\boldsymbol{f}$ refer to Earth's mass and external forces, respectively, and the asterisk (*) denotes parameters in the target frame. The linearized form of Eq. (35) in the target frame, known as the Clohessy-Wiltshire

(CW) equations, is expressed as [19]:

$$\ddot{x} - 2n\dot{z} = f_x \quad (36)$$
$$\ddot{y} + n^2 y = f_y \quad (37)$$
$$\ddot{z} + 2n\dot{x} - 2n^2 z = f_z \quad (38)$$

where:

$$n = \sqrt{\frac{\mu_{earth}}{R_o^3}} \quad (39)$$

Here, $R_o$ represents the orbital radius of the target spacecraft, and $n$ is the mean motion. To design the LQR controller, we begin by defining the state and input vectors as $\mathbf{x} = [x, y, z, \dot{x}, \dot{y}, \dot{z}]^T$, and $\mathbf{u} = [f_x, f_y, f_z]$, respectively. Subsequently, we derive the state space form of CW equations, expressed as:

$$\dot{\mathbf{x}} = A\mathbf{x} + B\mathbf{u} \quad (40)$$

where:

$$A = \begin{bmatrix} 0 & 0 & 0 & 1 & 0 & 0 \\ 0 & 0 & 0 & 0 & 1 & 0 \\ 0 & 0 & 0 & 0 & 0 & 1 \\ 0 & 0 & 0 & 0 & 0 & 2n \\ 0 & 0 & 0 & 0 & -n^2 & 0 \\ 0 & 0 & 0 & -2n & 0 & 2n^2 \end{bmatrix}; B = \begin{bmatrix} 0 & 0 & 0 \\ 0 & 0 & 0 \\ 0 & 0 & 0 \\ 1 & 0 & 0 \\ 0 & 1 & 0 \\ 0 & 0 & 1 \end{bmatrix} \quad (41)$$

In general, for the controllable pair of $(A, B)$, the control law for the LQR controller is given by [26]:

$$\mathbf{u} = -K_{LQR}\hat{\mathbf{x}} \quad (42)$$

Here, the symbol $\hat{\phantom{x}}$, denotes an estimated parameter. The controller gain $K_{LQR}$ is designed to minimize the following cost function:

$$J_c = \int_0^\infty (\mathbf{x}^T Q_c \mathbf{x} + \mathbf{u}^T R_c \mathbf{u}) dt \quad (43)$$

The weight matrices $Q_c$ and $R_c$ are determined through trial and error, with conditions $Q_c > 0$ and $R_c \geq 0$ satisfied. The controller gain $K_{LQR}$ is calculated using the following equation:

$$K_{LQR} = R^{-1} B^T S \quad (44)$$

where $S$ is the unique positive semidefinite solution of the algebraic Riccati equation:

$$A^T S + SA - SBR^{-1}B^T S + Q = 0 \quad (45)$$

It is important to note that due to the linearity and time-invariance of the considered system (LTI), the gain matrix $K_{LQR}$ is computed offline and does not require updating during the maneuver. Moreover, based on the separation principle of linear systems theory, the obtained gain can be incorporated into our presented network without imposing any condition on the estimator. The simulations in this section are conducted using the numerical values provided in TABLE 2, with a time duration of 360 seconds and a time step of 0.1. Additionally, the decoding matrices $D$ and $\bar{D}$ are defined using random samples from zero-mean Gaussian distributions with covariances of 1/50, and 1/2500, respectively.

TABLE 2
PARAMETERS FOR SATELLITE RENDEZVOUS

| Parameter | Value |
|---|---|
| $r_0$ (m) | $[70, 30, -5]^T$ |
| $v_0$ (m/s) | $[-1.7, -0.9, 0.25]^T$ |
| $x_0$ | $[r_0, v_0]^T$ |
| $\hat{x}_0$ | $x_0$ |
| $Q_c$ | $(1e-6)I_6$ |
| $R_c$ | $I_3$ |
| $Q$ | $(1e-12)I_6$ |
| $R$ | $(1e-2)I_2$ |
| $N$ | 350 |
| $\lambda$ | 0.001 |
| $\mu$ | 1 |
| $\nu$ | 0.0001 |
| $\delta_{MSIF}$ | 0.005 |

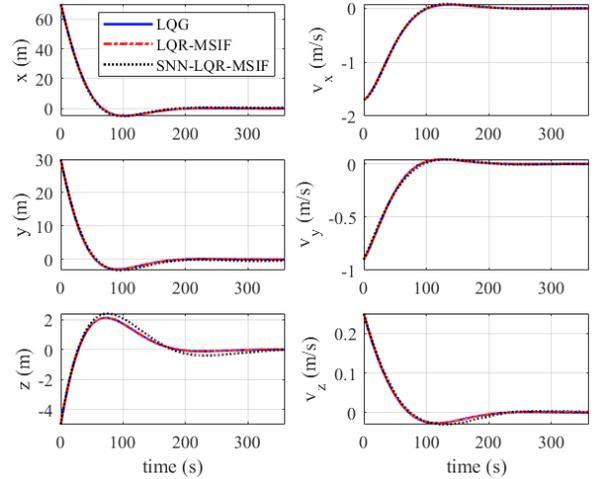

Fig. 11. Controlled states for satellite rendezvous obtained from various frameworks in normal condition.

Fig. 11 presents a comparison between SNN-LQR-MSIF and non-spiking LQG and LQR-MSIF in the context of the rendezvous maneuver problem. Each element of the system's state vector is individually compared. The results demonstrate that all considered frameworks successfully control the states, with errors smoothly converging to zero. Moreover, it is evident that the proposed SNN-based framework exhibits similar performance in controlling the states, aligning with the results obtained from the optimal non-spiking framework LQG. Notably, for states $z$, and $v_z$, some discrepancies are observed. For state $z$, the SNN-LQR-MSIF exhibits a slightly greater overshoot compared to non-spiking LQG and LQR-MSIF, but ultimately successfully controls the state error to zero. Furthermore, for state $v_z$ the result from SNN-LQR-MSIF exhibits minor deviation from non-spiking frameworks between $t = 100s$ and $t = 200s$. To provide quantitative insight into this comparison, averaged errors obtained from different methods after $t = 300s$ are presented in TABLE 3.

The results reveal that non-spiking methods deliver consistent accuracy, and the SNN-based method demonstrates acceptable accuracy. In summary, compared to traditional non-spiking frameworks like LQG and LQR-MSIF, the achieved results for controlled states affirm the acceptable performance of SNN-LQR-MSIF for the problem of satellite rendezvous, a critical maneuver in space robotic applications.

TABLE 3
AVERAGED ERROR FOR DIFFERENT METHODS

| State | KF | LQR-MSIF | SNN-LQR-MSIF |
|---|---|---|---|
| $x(m)$ | 0.0223 | 0.0222 | 0.3924 |
| $y(m)$ | 0.0057 | 0.0057 | 0.3626 |
| $z(m)$ | 0.0048 | 0.0048 | 0.0936 |
| $v_x(m/s)$ | 0.0012 | 0.0012 | 0.0018 |
| $v_y(m/s)$ | 0.0005 | 0.0005 | 0.0002 |
| $v_z(m/s)$ | 0.0005 | 0.0005 | 0.0030 |

To assess the computational efficiency of the SNN-based framework relative to conventional artificial neural networks (ANNs), we delve into the spiking pattern generated by the designed SNN, as showcased in Fig 12(a). This vividly illustrates the network's efficient execution of its task. Upon closer examination, as depicted in Fig. 12(b), during the initial 2000 time-steps (before $t = 100$s), when state-vector errors are sizable, the network exhibits heightened neural activity, with approximately 20% of neurons being active. Subsequently, the population of active neurons gently declines and remains relatively constant, with minor fluctuations hovering around 5% for the remainder of the simulation. In essence, the network accomplishes its task while utilizing a mere 2.4% of possible spikes over the entire simulation duration, in stark contrast to traditional ANNs that consume 100% of potential spikes. This underscores the computational efficiency of SNN-LQR-MSIF in simultaneously handling estimation and control for satellite rendezvous. Moving on to assess the robustness of the SNN-LQR-MSIF against modeling uncertainties, we introduce a 10% error into the dynamic transition matrix $\hat{A} = 0.9A$ used within the framework. Fig. 13 demonstrates the results for controlled states using aforementioned strategies. This figure underscores that SNN-LQR-MSIF exhibits higher sensitivity to modeling uncertainties compared to non-spiking strategies. However, it also presents that SNN-LQR-MSIF effectively control the system, with all the errors gracefully converging to zero. Furthermore, TABLE 4 presents averaged errors obtained from controlled states after $t = 300$s, reaffirming the findings depicted in Fig. 13. To further evaluate the robustness of SNN-LQR-MSIF against external disturbances, such as instability in the working environment, we introduce measurement outliers. This scenario is configured so that unmodeled measurement outliers are injected into the system at $t = 100$s, $t = 150$s, and $t = 200$s. Notably, to introduce the outliers at these time steps, the measurement system noise is scaled by a factor of 200. Fig. 14 illustrates the results for various frameworks in this scenario. Similar to modeling uncertainties, it reveals that the SNN-LQR-MSIF is more sensitive to measurement outliers compared to non-spiking strategies. However, it effectively maintains control, with all errors converging to zero. Corresponding averaged errors from the controlled states after $t = 300$s is presented in TABLE 5, thus reinforcing the insights gleaned from the data depicted in Fig. 14.

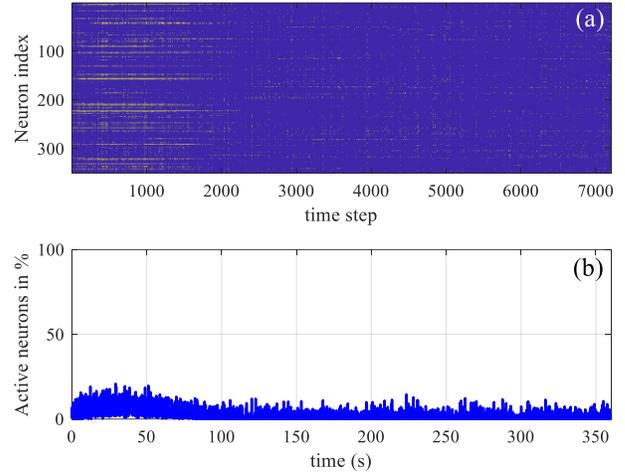

Fig. 12. Spiking pattern and temporal variation of active neurons population obtained from SNN-LQR-MSIF for satellite rendezvous maneuver, (a) spiking pattern, (b) temporal variation of active neurons.

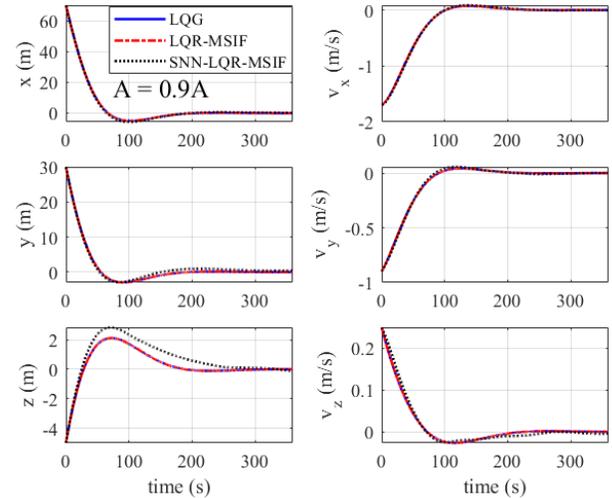

Fig. 13. Controlled states for satellite rendezvous maneuver obtained from various frameworks for uncertain model.

TABLE 4
AVERAGED ERROR FOR DIFFERENT METHODS – UNCERTAIN MODEL

| State | LQG | LQR-MSIF | SNN-LQR-MSIF |
|---|---|---|---|
| $x(m)$ | 0.0223 | 0.0222 | 0.3059 |
| $y(m)$ | 0.0058 | 0.0057 | 0.4001 |
| $z(m)$ | 0.0049 | 0.0049 | 0.0082 |
| $v_x(m/s)$ | 0.0012 | 0.0012 | 0.0030 |
| $v_y(m/s)$ | 0.0005 | 0.0005 | 0.0001 |
| $v_z(m/s)$ | 0.0005 | 0.0005 | 0.0035 |

Fig. 15 provides insight into the spiking pattern of SNN-LQR-MSIF in the presence of measurement outliers. In Fig. 15(a), the network reacts to disturbances by increasing the number of active neurons, rapidly rejecting disturbances in just 2-3 time steps. Fig. 15(b) quantifies this by depicting the variation in the population of active neurons in percentage terms. The figure highlights a significant increase in the proportion of active neurons, rising from approximately 10% to nearly 50%.

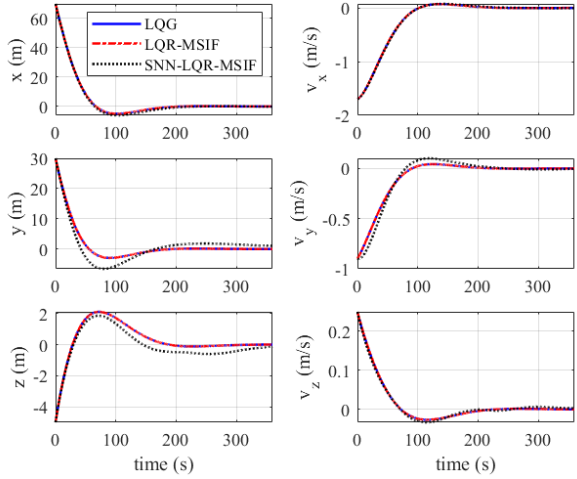

Fig. 14. Controlled states for satellite rendezvous maneuver obtained from various frameworks subjected to measurement outlier.

TABLE 5
AVERAGED ERROR FOR DIFFERENT METHODS – MEASUREMENT OUTLIER

| State | LQG | LQR-MSIF | SNN-LQR-MSIF |
|---|---|---|---|
| $x(m)$ | 0.0223 | 0.0222 | 0.0001 |
| $y(m)$ | 0.0058 | 0.0057 | 1.2319 |
| $z(m)$ | 0.0049 | 0.0049 | 0.2518 |
| $v_x(m/s)$ | 0.0012 | 0.0012 | 0.0050 |
| $v_y(m/s)$ | 0.0005 | 0.0005 | 0.0077 |
| $v_z(m/s)$ | 0.0005 | 0.0005 | 0.0047 |

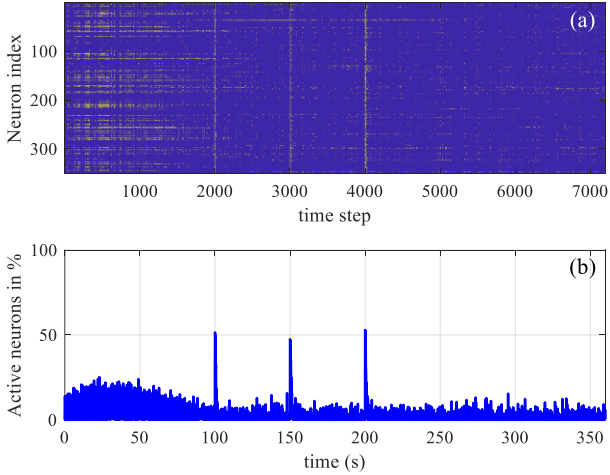

Fig. 15. Spiking pattern and temporal variation of active neurons population obtained from SNN-LQR-MSIF for satellite rendezvous maneuver subjected to measurement outlier, (a) spiking pattern, (b) temporal variation of active neurons.

Finally, the results obtained in this section affirm that the framework proposed in this study demonstrates computational efficiency for such problems. Compared to traditional computing strategies like LQR-MSIF and LQG, it exhibits good and comparable performance in terms of robustness and accuracy.

## V. CONCLUSION

In this presented study, we delved into the crucial challenges of concurrent estimation and control within dynamical systems, underscoring its paramount importance. As the complexity and safety considerations associated with mission-critical tasks continue to intensify, the demand for computationally efficient and dependable strategies has become increasingly imperative. Moreover, in the real-world application landscape, rife with uncertainties such as environmental instability, external disturbances, external disturbances, and unmodeled dynamics, the call for robust solutions capable of navigating these challenges is resounding. To answer this call, we introduced a novel approach grounded in biologically plausible principles. Our framework harnessed the potential of a recurrent spiking neural network (SNN), composed of leaky integrate-and-fire neurons, bearing resemblance to a linear quadratic regulator (LQR) enriched by the insights of a modified sliding innovation filter (MSIF). This innovative amalgamation endowed the SNN-LQR-MSIF with the robustness inherited from the MSIF, while concurrently infusing it with computational efficiency and scalability inherent in SNNs. Importantly, the elimination of the need for extensive training, owing to spike coding theories, empowered the design of SNN weight matrices grounded in the dynamic model of the target system.

In the face of a diverse array of uncertainties, including modeling imprecision, unmodeled measurement outliers, and occasional neuron silencing, we conducted a thorough comparative analysis. The SNN-LQR-MSIF underwent meticulous evaluation, alongside its non-spiking counterpart, the LQR-MSIF, and the well-established optimal approach, linear quadratic Gaussian (LQG). This evaluation spanned both linear benchmark problems and the satellite rendezvous maneuver, a mission-critical task within the realm of space robotics. The results of our investigation underscored the SNN-LQR-MSIF's commendable performance. It demonstrated competitive advantages in terms of computational efficiency, reliability, and accuracy, positioning it as a promising solution for addressing concurrent estimation and control challenges. Looking forward, we envisage the development of learning-based concurrent robust estimation and control frameworks, leveraging the capabilities of SNNs and predictive coding. These endeavors represent exciting prospects for future research in this domain, further enhancing the state-of-the-art in dynamical system control and estimation.

## VI. CONFLICT OF INTEREST STATEMENT

The authors declare that they do not possess any conflicts of interest pertinent related to this research. This study was executed with the utmost objectivity and impartiality, and the results articulated herein stem from a meticulous and unbiased scrutiny and comprehension of the data. The authors maintain that they harbor no financial or personal affiliations with individuals or entities that could conceivably introduce bias into the findings or exert influence over the conclusions drawn from this study.